\pdfoutput=1

\documentclass[11pt]{article}

\usepackage[final]{acl}
\usepackage{tabularx}
\usepackage{algorithm}
\usepackage{algpseudocode}
\usepackage{array}

\usepackage{times}
\usepackage{latexsym}
\usepackage{booktabs}
\usepackage{amsmath,amsfonts,amssymb}
\usepackage{graphicx}
\usepackage{float}
\usepackage{url}
\usepackage[T1]{fontenc}

\usepackage[utf8]{inputenc}

\usepackage{microtype}

\usepackage{inconsolata}
\usepackage{babel}
\usepackage{graphicx}
\title{Sharif-MGTD at SemEval-2024 Task 8: A Transformer-Based Approach to Detect Machine Generated Text}

\author{
    \textbf{Seyedeh Fatemeh Ebrahimi}$^\clubsuit$, \textbf{Karim Akhavan Azari}$^\clubsuit$, \textbf{Amirmasoud Iravani}$^\bigstar$ \\
    \textbf{Arian Qazvini}$^\lozenge$, \textbf{Pouya Sadeghi}$^\dag$, \textbf{Zeinab Sadat Taghavi}$^\clubsuit$, \textbf{Hossein Sameti}$^\clubsuit$ \\
    Ferdowsi University of Mashhad, Mashhad, Iran$^\bigstar$ \\
    Amirkabir University of Technology, Tehran, Iran$^\lozenge$ \\
    University of Tehran, Tehran, Iran$^\dag$ \\
    Sharif University of Technology, Tehran, Iran$^\clubsuit$ \\
    \\
    \texttt{\{sfati.ebrahimi, karim.akhavan, zeinabtaghavi, sameti\}@sharif.edu} \\
    \texttt{a.iravani@mail.um.ac.ir} \\
    \texttt{a.qazvini@aut.ac.ir} \\
    \texttt{pouya.sadeghi@ut.ac.ir}
}

\begin{document}
\maketitle
\begin{abstract}
Detecting Machine-Generated Text (MGT) has emerged as a significant area of study within Natural Language Processing. While language models generate text, they often leave discernible traces, which can be scrutinized using either traditional feature-based methods or more advanced neural language models. In this research, we explore the effectiveness of fine-tuning a RoBERTa-base transformer, a powerful neural architecture, to address MGT detection as a binary classification task. Focusing specifically on Subtask A (Monolingual - English) within the SemEval-2024 competition framework\footnote{\url{https://semeval.github.io/SemEval2024/}}, our proposed system achieves an accuracy of 78.9\% on the test dataset, positioning us at 57th among participants. Our study addresses this challenge while considering the limited hardware resources, resulting in a system that excels at identifying human-written texts but encounters challenges in accurately discerning MGTs.
\end{abstract}

\section{Introduction}

Recent advancements in large language models (LLMs) have endowed them with an impressive capability to generate written text that closely resembles human writing \citep{Adelani2019GeneratingSF, Radford2019LanguageMA}. However, this technological progress brings along significant challenges, as the proliferation MGT poses various threats in digital environments. MGTs have been implicated in spreading misinformation in online reviews, eroding public trust in political or commercial campaigns, and even facilitating academic fraud \citep{Crothers2022MachineGeneratedTA, Song2015CrowdTargetTD, Tang2023TheSO}.
The identification of MGT remains a pressing concern, as distinguishing between human-written and machine-generated content is often challenging for humans. Consequently, there is a growing imperative to develop automatic systems capable of discerning MGT \citep{Mitchell2023DetectGPTZM}. In this study, we address this challenge within the English language context using the dataset provided by \citet{Wang2023M4MM}.\newline

As highlighted in \citet{wang-EtAl:2024:SemEval20245} overview paper on the task, recent approaches to MGT detection predominantly employ binary classification methods. Existing literature highlights the superior performance of transformer-based methods over alternative approaches \citet{Wang2024M4GTBenchEB}. However, a significant challenge in utilizing these models lies in the requirement for GPU hardware and computational resources. Our study aims to address this challenge within the constraints of limited hardware capacity. Keeping this in mind, we propose a system that leverages fine-tuning of the RoBERTa transformer model \citep{Liu2019RoBERTaAR} to automatically classify input text as either human-written or machine-generated. Our system architecture involves augmenting the RoBERTa-base model with a Classifier Head. The Embeddings component facilitates contextual understanding of texts, while the Encoder component processes input texts in parallel, and the Classifier Head performs binary classification by linearly outputting a single value.\newline

Our proposed system achieves an accuracy of 78.9\% on the test data, surpassing the average results provided by the task's baseline and ranking 57th among 140 participants. The area under the ROC curve (AUC) metric is measured at 0.69. While the ROC curve analysis demonstrates our model's capability to classify substantial portions of positive cases, its proximity to the diagonal line indicates room for further improvement. Notably, our primary challenge stemmed from computational constraints, which limited our ability to implement larger token sizes or batch sizes. Further discussions reveal that our system encounters difficulties in accurately detecting MGTs. To facilitate reproducibility and further research in this area, the code for our system is available on GitHub\footnote{\url{https://github.com/Sharif-SLPL/Sharif-MGTD}}.

\section{Background}

\subsection{Dataset Overview}

SemEval-2024 Task 8 \citep{wang-EtAl:2024:SemEval20245} comprises three subtasks, with our investigation centering on Subtask A: binary classification of human-written versus MGT. Specifically, we concentrated our efforts on analyzing English monolingual data, as outlined dataset is provided by \citet{Wang2023M4MM}. 

Subtask A encompasses a dataset consisting of 119,757 training examples and 5,000 development examples, all presented in JSON format. Each data instance includes the following attributes: 
\begin{itemize}
    \item \textit{id}: An identifier number for the example.
    \item \textit{label}: A binary label indicating whether the text is human-written (0) or machine-generated (1).
    \item \textit{text}: The actual textual content.
    \item \textit{model}: The AI machine responsible for generating the text.
    \item \textit{source}: The web domain from which the text originates.
\end{itemize}

\subsection{Related Work}

MGT detection is feasible through both traditional feature-based methods and neural language models. \citet{Frhling2021FeaturebasedDO} and \citet{NguyenSon2018IdentifyingCT} discussed how feature-based methods leverage statistical techniques. These methods primarily utilize frequency features such as TF-IDF, linguistic cues, and text style \citep{Frhling2021FeaturebasedDO}. However, feature-based methods have limitations, as different samplings in language models can lead to varied generated outputs \citep{Holtzman2019TheCC}.
In contrast, methods that harness neural language models, particularly those employing transformer models, have shown high effectiveness \citep{Crothers2022MachineGeneratedTA}. Neural language model methods often involve zero-shot classification or fine-tuning pre-trained language models \citep{Sadasivan2023CanAT}. Grover by \citet{Zellers2019DefendingAN}, RankGen by \citet{Krishna2022RankGenIT}, and DetectGPT \citep{Mitchell2023DetectGPTZM} are prominent examples of zero-shot methods. However, these methods may be misleading at times and exhibit limited performance in out-of-domain tasks \citep{Crothers2022MachineGeneratedTA, Wang2023M4MM}.\newline

\citet{Bakhtin2019RealOF} demonstrated outstanding performance in MGT detection by harnessing bidirectional transformers. Additionally, \citet{Solaiman2019ReleaseSA} highlight that the zero-shot methods often fall short compared to a simple TF-IDF baseline when detecting texts from diverse domains. He argues that bidirectional transformers offer significant advantages for MGT detection, advocating for the fine-tuning of these models as a superior alternative to zero-shot methods. In this regard, \citet{Rodriguez2022CrossDomainDO} observed a significant enhancement in performance of cross-domain MGT detection by fine-tuning the RoBERTa detector. \newline

\citet{Jawahar2020AutomaticDO} conducted a comprehensive survey of various approaches to developing MGT detectors. Their findings suggest that fine-tuning the RoBERTa detector consistently delivers robust performance across diverse MGT detection tasks, surpassing the efficacy of traditional machine learning models and neural networks. Additionally, \citet{Crothers2022MachineGeneratedTA} reported a notable trend towards the increased utilization of bidirectional transformer architectures, particularly RoBERTa, in MGT detection tasks. Lastly, \citet{Wang2024M4GTBenchEB} conducted a comprehensive benchmark of supervised methods on M4 dataset. Their findings revealed that transformer models such as RoBERTa and XLM-R exhibited superior performance across all tests, respectively achieving 99.26\% and 96.31\% accuracy in MGT binary classification.  \newline

While this review does not provide a comprehensive examination of all aspects of MGT detection, prior research underscores the prevalence of transformer-base methods, like RoBERTa and XLM-R, in comparison to alternative approaches, especially in supervised tasks. Moreover, the superiority of RoBERTa over other models is evident. A significant challenge for studies utilizing pre-trained transformer models lies in the necessity for robust GPU hardware and computational resources. 
\section{System Overview}
This section presents an overview of our system's architecture, highlighting implementation details and challenges. Drawing on the preceding works discussed above, which showed the efficacy of fine-tuning RoBERTa models, our system aims to attain peak performance in MGT detection while optimizing configurations for limited hardware resources.  \newline
The decision to employ the transformer architecture for detecting synthetic texts is motivated by its capacity to capture intricate dependencies within textual data. This choice seems logical considering that such texts often exhibit semantic features that can be harnessed for fact-checking, cohesion, coherence, and other properties that may unveil their origin (\citet{raj-etal-2020-solomon}. In contrast to traditional architectures, the transformer model overcomes the constraints of fixed window sizes or sequential processing, enabling it to utilize contextual information from the entire input sequence. Additionally, the self-attention mechanism empowers the model to selectively focus on pertinent segments of the input, rendering it highly effective for tasks necessitating long-range dependencies and contextual comprehension. \newline
As for RoBERTa, it is specifically chosen for its extensive training duration, broader dataset coverage, ability to handle longer sequences, and focus on Natural Language Understanding tasks, making it more suitable than other BERT-based models. Additionally, a wealth of research, such as the recent study of \citet{Wang2024M4GTBenchEB}, has further highlighted the inherent potential of RoBERTa for this specific task.

\subsection{Core Algorithms and System Architecture}

At the core of our system lies the concept of binary classification, distinguishing input texts as either machine-generated or human-written through fine-tuning a pre-trained RoBERTa transformer \citep{Liu2019RoBERTaAR}. Our system architecture entails augmenting the RoBERTa-base model with a Classifier Head. The RoBERTa model's Embeddings component incorporates a 768-dimensional embedding matrix, alongside position and token type embeddings, enhancing contextual understanding. The Encoding component features a 12-layer RoBERTaEncoder, each layer employing a multi-head self-attention mechanism. This facilitates simultaneous attention to different parts of the input text, crucial for analyzing textual similarities. Intermediate sub-layers utilize a fully connected feed-forward network with GELU activation, followed by an output sub-layer for feature transformation and normalization. 

The Classifier Head, integrated into the Encoder for sequence classification, comprises a linear layer with 768 input features and a dropout layer to mitigate over-fitting. The final output is generated through an additional linear layer with a solitary output neuron, making it conducive to binary classification tasks. In essence, the primary model processes input data, with the Classifier Head making predictions. When viewed as a regression task, the Classifier produces a linear output tailored for a singular class, providing a probabilistic value.
Implementation of the system is facilitated using PyTorch, incorporating specific parameters such as the AdamW optimizer \citep{Radford2018ImprovingLU} and the CrossEntropyLoss function \citep{Hui2020EvaluationON}. AdamW, renowned for training deep neural networks, integrates weight decay to mitigate over-fitting. The Cross Entropy Loss function, commonly employed in multi-class classification scenarios, combines softmax activation with negative log-likelihood loss. The training process involves iterating through the entire dataset for two epochs, with early stopping mechanisms in place to terminate training at the optimal point.

\subsection{System Challenges}
While larger machine-generated documents often exhibit more discernible patterns and clues, such as incoherence or repetition, they also entail substantial computational costs. Our primary challenge lay in efficiently processing these large documents using cost-effective computing systems. To mitigate this challenge, we explored strategies such as reducing token size and batch size. However, these adjustments necessitate trade-offs, potentially leading to reduced accuracy or increased processing time.

Our system was trained using a token size of 512, but optimal performance could potentially be achieved with larger token sizes, such as 1024 or 2048, given sufficient computing resources.

\section{Experimental Setup}
\subsection{Dataset}
Table 1 presents detailed statistics on the dataset used for each class. 

\begin{table}[H]
\resizebox{\columnwidth}{!}{%
\centering
\begin{tabular}{l|cccc}
\hline
\textbf{Class/Split} & \textbf{Train} & \textbf{Test} & \textbf{Development} \\
\hline
Human-Written Text & 57075 & 6276 & 2500 \\
\hline Machine-Generated Text & 50706 & 5700 & 2500 \\
\hline
\end{tabular}%
}
\caption{Dataset Statistics}
\end{table}
As shown in Table 1, nearly 90\% of the dataset is dedicated to training, while the remainder is used for evaluation. To enhance model performance, we utilized the entire development dataset for model selection, compensating for the scarcity of training data.

\subsection{Pre-processing and Hyper-Parameter Tuning}
Input texts are tokenized using the RoBERTa tokenizer before processing, both during training and inference. Our hyper-parameter tuning process involved a comprehensive exploration across various parameter ranges. Specifically, we conducted experiments with learning rates ranging from 0.0001 to 0.00004, dropout rates spanning from 0.1 to 0.3, batch sizes varying between 4 and 16, and token sizes ranging from 64 to 1024. Through experimentation and analysis, we determined the optimal hyper-parameter settings, which are as follows: a learning rate of 0.00004, a dropout rate of 0.1, a token size of 512, a batch size of 10, and a weight decay of 0.01. Further details are given in Appendix A.

As illustrated in Appendix A, the number of training instances is correlated with the input token size and may influence the model accuracy. Given the length of input texts, a suitable token size is essential to capture all tokens adequately. However, computational costs associated with larger token sizes present a significant challenge during model training. Consequently, we selected 512 as the optimal token size. Truncation was employed during tokenization to accommodate the chosen token size, ensuring efficient model training without compromising data representativeness.

\subsection{Training Procedure}
For training the model, we utilized the Task dataset \citet{Wang2023M4MM}, which underwent preprocessing by tokenizing the text into sub-word units and padding sequences to a fixed length. CrossEntropyLoss was employed as the loss function. The implementation also involved the AdamW optimizer, known for its effectiveness in training deep neural networks and its incorporation of weight decay to address over-fitting. The Adam optimizer was utilized with a learning rate of 4e-05. During training, the loss was monitored on a held-out validation set, and early stopping was applied to prevent over-fitting. Early stopping was implemented with the condition that the training loss reached a specific threshold (0.35 in this case), typically occurring around the third epoch. Therefore, if there was no improvement in the validation loss for a certain number of epochs, training was halted to prevent over-fitting of the model.

\subsection{Evaluation Measures}
The evaluation of our model involves calculating its accuracy in predicting whether a text is human-written or machine-generated. Accuracy, a fundamental metric in classification tasks, assesses the overall correctness of predictions and is calculated as:

\begin{equation}
\begin{aligned}
Accuracy = \frac{ni}{N} \times 100
\end{aligned}
\end{equation}
where \( n_i \) represents the number of correctly classified instances, and N is the total number of instances.

\section{Results}

Using the official accuracy metric of SemEval-2024 Task 8 \citep{wang-EtAl:2024:SemEval20245}, our system achieved the following accuracy scores on different data splits:

\begin{table}[H]
\centering
\begin{tabular}{l|cccc}\
\textbf{Language /Split} & \textbf{Devset} & \textbf{Testset} \\
\hline
English & 74.8\% & 78.9\% \\
\hline
\end{tabular}
\caption{Accuracy Metric}
\end{table}

A direct comparison of our results with prior works is challenging due to the unique nature of our research. To the best of researchers’ knowledge, the most comprehensive benchmark on supervised MGT detection is presented by \citet{Wang2024M4GTBenchEB} using the M4 dataset and employing RoBERTa, XLM-R, GLTR-LR, GLTRSVM, Stylistic-SVM, and NELA-SVM. However, our primary objective was to determine strategies for addressing limited hardware resources as discussed in Appendix A.

As a contribution to this field, through repeated experiments, we identified that among hyperparameters, token size plays a slightly more significant role in model accuracy. While the system's accuracy is influenced by increasing the token size, drawing meaningful scientific conclusions necessitates further controlled experiments. Additionally, the expansion of token size is restricted by hardware limitations, requiring a detailed investigation with robust computational resources like GPU or TPU. Considering the constraints of Google Colab's\footnote{\url{https://colab.research.google.com}} Free runtimes, we opted for a token size of 512 as a balance between hardware limitations and time constraints. Consequently, based on the official accuracy metric of SemEval2024 Task 8 \citep{wang-EtAl:2024:SemEval20245}, our system achieved the following accuracy scores on various data splits:

\begin{figure}[h]
  \centering
  \includegraphics[width=0.9\linewidth]{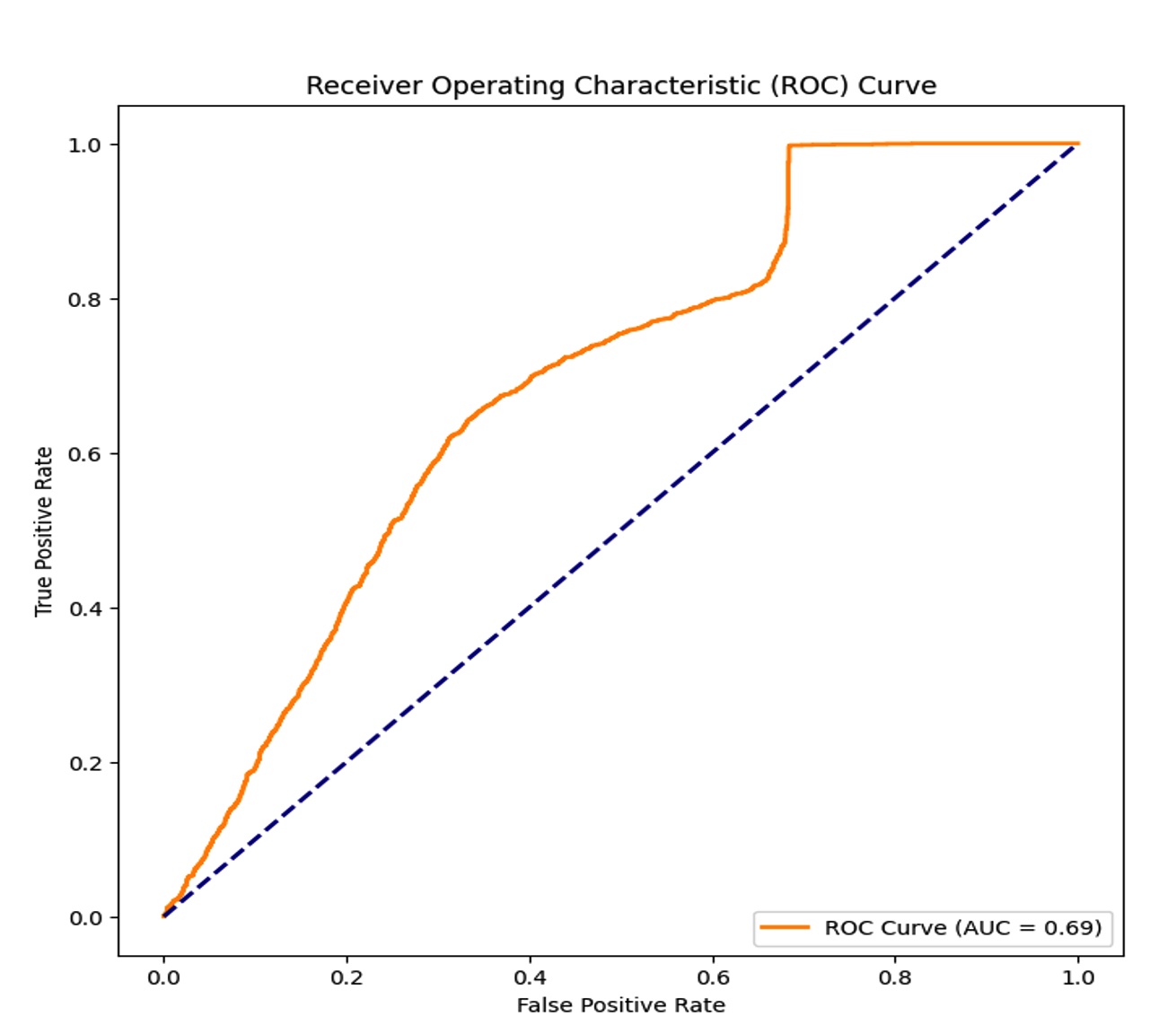}
  \caption{ The ROC Curve Plot}
\end{figure}

\begin{figure}[h]
  \centering
  \includegraphics[width=0.9\linewidth]{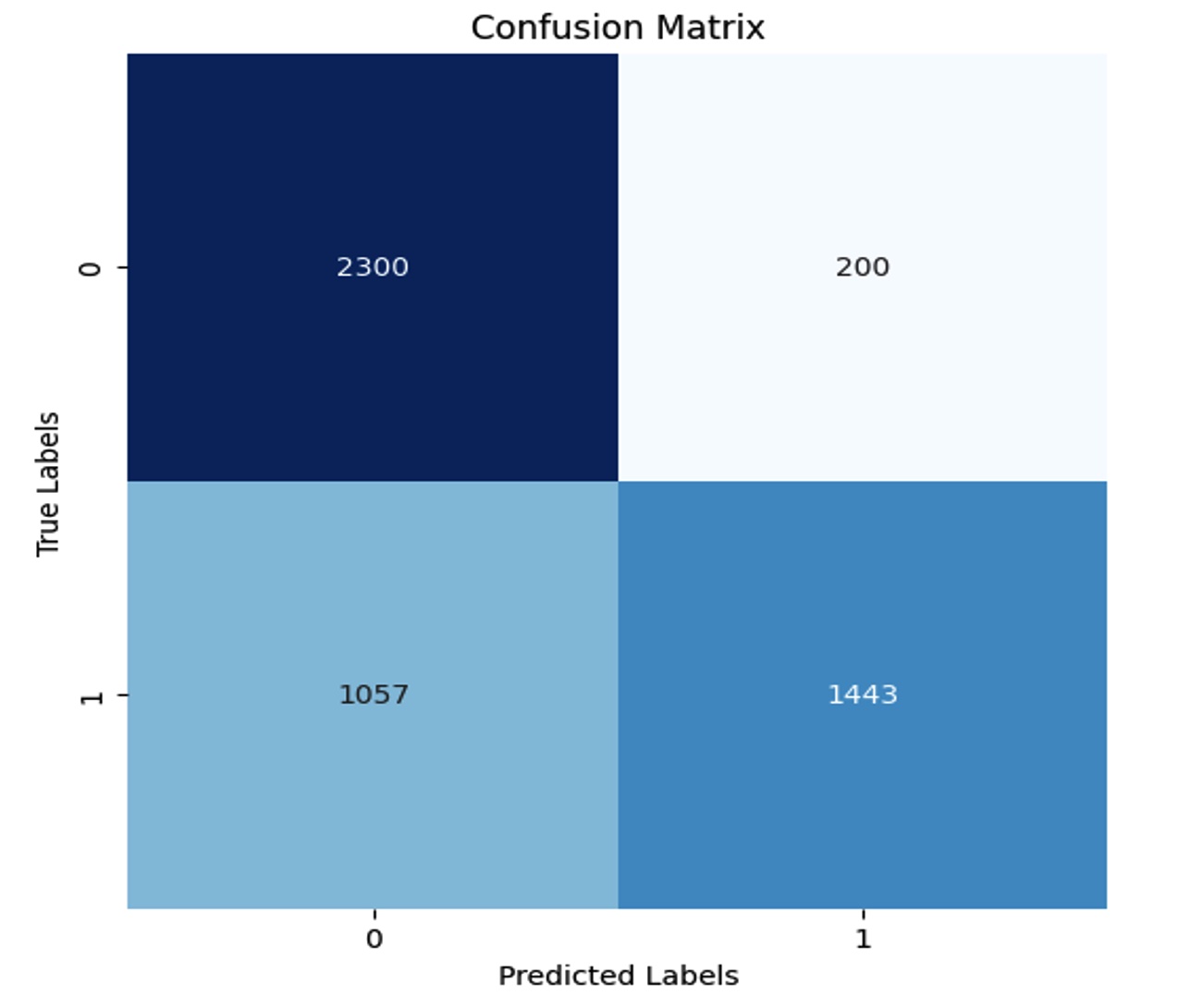}
  \caption{ The Confusion Matrix Plot}
\end{figure}
The evaluation of our model also included analysis of the Area Under the Curve (AUC), a crucial metric that reflects the discriminative power of a binary classification model. Our fine-tuned RoBERTa model demonstrated an AUC of 0.69, suggesting its ability to effectively distinguish between positive and negative instances. Figure 1 illustrates the Receiver Operating Characteristic (ROC) Curve, depicting the model's capability to accurately classify a significant proportion of positive cases. However, the proximity of the curve to the diagonal line suggests opportunities for further enhancement.

Interestingly, analysis of the confusion matrix, as depicted in Figure 2, revealed notable patterns in our model's classification tendencies. While our system effectively identified human-written documents with low False Positives, it exhibited difficulties in correctly identifying MGTs. This observation suggests potential areas for refinement, particularly in enhancing the model's ability to detect subtle cues and characteristics unique to machine-generated content.

Overall, our study contributes to the ongoing efforts in the field of NLP by showcasing the effectiveness of fine-tuned transformer models, particularly RoBERTa, in MGT detection tasks. Moving forward, future research directions could explore novel approaches to mitigate computational costs and further improve the performance of MGT detection systems, ultimately advancing the capabilities of NLU models in real-world applications.

\section{Conclusion}

In summary, our study focused on fine-tuning a RoBERTa-base transformer model for binary classification, specifically in distinguishing human-written from MGT. While our system showed promise in identifying human-written text, it faced challenges with accurately classifying machine-generated content. As discussed in Appendices A and B, we recommend exploring larger token sizes to improve model performance, albeit with awareness of computational costs. Additionally, we advocate for the development of low-cost algorithms capable of efficient processing across hardware platforms. Our findings contribute to advancing MGT detection, with implications for combating misinformation and enhancing cyber-security in the digital age.

\section*{Acknowledgments}

We appreciate the Speech and Language Processing Laboratory at Sharif University of Technology\footnote{\url{https://github.com/Sharif-SLPL}} for providing us with this opportunity for collaborative work.

\bibliography{custom}
\clearpage
\appendix
\section{Hyper-Parameter Tuning}
\label{sec:appendix}

\begin{figure}[h]
  \centering
  \includegraphics[width=1\linewidth]{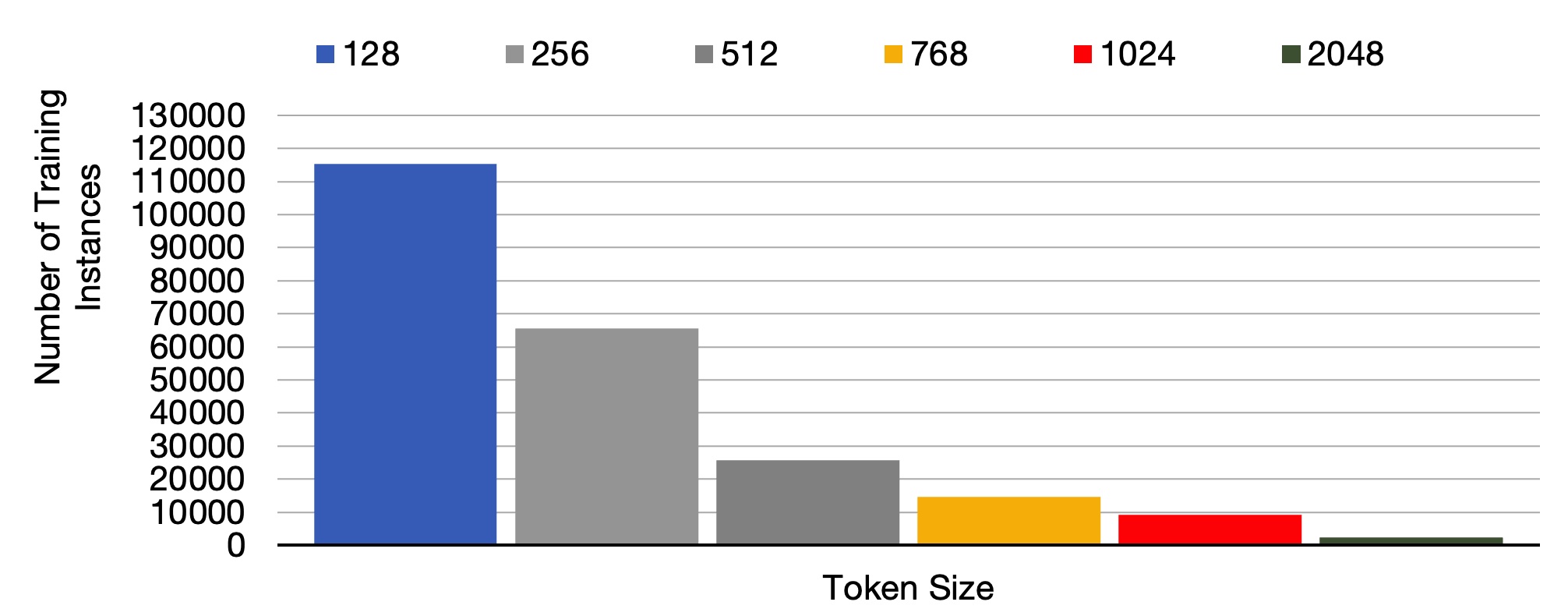}
  \caption{ Number of Training Instances by Token Size}
\end{figure}
To determine the appropriate settings for hyper-parameters, we utilized Google Colab's free GPU runtime. Free Colab users have access to GPU and TPU runtimes without charge for a maximum of 12 hours. The GPU runtime includes an NVIDIA Tesla K80 with 12GB of VRAM. [Date: 5 Dec 2023]. We were unable to use premium runtime accounts due to financial issues arising from Iran sanctions. Therefore, we couldn't change our model's token size to larger than 512 due to the 12-hour time limit in free Colab. To understand the impact of increasing token size, we aimed to experiment on a local laptop GPU.\newline

During the experiments aimed at finding the proper token size, we encountered the "CUDA error: device-side assert triggered" frequently, which was resolved by restarting the session. Our experiments were conducted using an RTX 2060 mobile with 6 GB of VRAM. Throughout all experiments, we maintained fixed parameters, including Number of Epochs = 3, Train Split = 0.7, and Learning Rate = 4e-05. Increasing the Max Length from 512 to 1024 in this experimental setup resulted in an improvement in Test Accuracy by at least 2\%. However, this enhancement came at the cost of a nearly 15-fold decrease in training speed, making it challenging to implement on limited hardware. Additionally, this requires plenty of controlled experiments by researchers to shed light on finding the proper hyper-parameters. \newline

\section{Detect-GPT as a Zero-Shot Method}
In our pursuit of effective MGT detection, we also experimented with \citet{Mitchell2023DetectGPTZM}Detect-GPT model, a zero-shot approach utilizing probability curvature analysis. Training the model resulted in an accuracy rate of 60\%, and when applied to a test dataset of approximately 1500 samples, it achieved a remarkable accuracy of approximately 84\%. We conducted a comprehensive analysis by implementing 10 perturbations for each dataset. To address data and mask filling tasks, we employed the T5 small model, leveraging its robust capabilities. Furthermore, to accurately assess the log likelihood, we utilized the GPT-2 model, ensuring precise calculations and reliable results.This method surpassed alternative text detection methodologies, demonstrating superior accuracy and reliability in identifying MGT. Notably, the inclusion of threshold configuration added granularity to the experiment, enabling fine-tuning of detection sensitivity across varying threshold settings.

\end{document}